\documentclass{article}
\usepackage{iclr2026_conference,times}

\usepackage{amsmath}
\usepackage{amssymb}
\usepackage{booktabs}
\usepackage{multirow}
\usepackage{longtable}
\usepackage{tabularx}
\usepackage{pdflscape}
\usepackage{float}
\usepackage{graphicx}
\usepackage{xcolor}
\usepackage{hyperref}
\usepackage{url}
\usepackage{xspace}
\usepackage{enumitem}
\usepackage{algorithm}
\usepackage{algpseudocode}
\usepackage{capt-of}

\usepackage[table]{xcolor}
\usepackage[normalem]{ulem}

\definecolor{oursbg}{gray}{0.93}

\definecolor{gaincolor}{RGB}{185,105,105}
\newcommand{\gain}[1]{%
  {\scriptsize\textcolor{gaincolor}{\bfseries\,($\uparrow$#1)}}%
}

\newcommand{\best}[1]{\textbf{#1}}
\newcommand{\second}[1]{\uline{#1}}

\usepackage{amsthm}

\newcolumntype{Y}{>{\raggedright\arraybackslash}X}
\newcolumntype{L}[1]{>{\raggedright\arraybackslash}p{#1}}

\definecolor{uclablue}{rgb}{0.15, 0.45, 0.68}
\hypersetup{breaklinks, colorlinks=true, citecolor=uclablue, linkcolor=uclablue, urlcolor=uclablue}

\title{Scalable Frequency- and Length-Aware \\Subdocument Deduplication for Large \\Language Model Pretraining}

\newcommand{\authorsep}{\hspace{2.2em}}

\author{%
\mbox{Hai Wang$^{*\dagger}$}\authorsep
\mbox{Chenhao Wang$^{*}$}\authorsep
\mbox{Qifeng Cai$^{*}$}\authorsep
\mbox{Yixiu Liu}\authorsep
\mbox{Miao Peng}
\\[0.18cm]
\mbox{\textbf{Nuo Chen}}\authorsep
\mbox{\textbf{Yuanlin Tu}}\authorsep
\mbox{\textbf{Chengcheng Xu}}\authorsep
\mbox{\textbf{Feng Zhang}}
\\[0.20cm]
Hunyuan Team, Tencent
}

\iclrfinalcopy
\begin{document}
\maketitle

% ==================================================================
\begin{abstract}
Large-scale pretraining corpora contain substantial duplicate content. Although document-level deduplication is widely used, removing subdocument-level redundancy remains challenging. At corpus scale, suffix-array-based methods are commonly applied independently within shards, leaving cross-shard duplicates undetected and making the resulting retention behavior sensitive to the sharding configuration. Hash-based methods enable global exact duplicate counting, but often rely on fixed copy-retention policies that cannot accommodate heterogeneous repetition patterns. We propose a scalable subdocument deduplication framework that decouples duplicate detection from copy retention. It identifies duplicate groups through natural-boundary segmentation, normalized exact hashing, and distributed aggregation, and then applies an explicit frequency- and length-aware retention policy that allocates an adaptive copy budget to each group, retaining more copies of low-frequency or short repetitions while more aggressively deleting high-frequency or long ones. Experiments on FineWeb-Edu and a code-containing web corpus show that models trained on data processed by our method achieve the best overall performance among the evaluated settings. These results underscore the importance of explicit copy-retention control.
\end{abstract}

\begingroup
\renewcommand{\thefootnote}{}
\footnotetext{
  $^{*}$Equal contribution.\qquad
  $^{\dagger}$Project leader.
}
\endgroup

\setcounter{footnote}{0}
\renewcommand{\thefootnote}{\arabic{footnote}}

\section{Introduction}
\label{sec:introduction}

Pretraining is the core stage during which large language models acquire general knowledge and foundational capabilities, and its effectiveness depends not only on the scale of training data but also on data quality~\citep{hoffmann2022chinchilla,li2024dclm,penedo2024fineweb}. However, large-scale pretraining data often contain substantial duplicate content, such as web templates, announcements, quoted text, copied code, and repeated segments arising from different versions~\citep{dodge2021documenting,lee2022deduplicating,kohlschuetter2010boilerplate}. Such duplication wastes limited training compute and can cause a disproportionate degradation in model performance that may not be fully offset by training on additional tokens~\citep{hernandez2022scaling}. Consequently, deduplication has become a critical step in the data preprocessing pipeline for large language model pretraining.

Most existing pretraining pipelines include document-level deduplication, typically using exact hashing or MinHash-based near-duplicate detection~\citep{broder1997resemblance,gao2020pile,penedo2024fineweb,grattafiori2024llama3}. These methods are efficient and scalable, but treat the entire document as the unit of deduplication. In real-world corpora, however, repeated content often covers only part of a document~\citep{lee2022deduplicating}. For example, documents with otherwise distinct content may share web templates, navigation bars, copyright notices, quoted passages, or code snippets. Lowering the document-similarity threshold may detect more such local overlap, but also risks removing documents that contain only limited shared text alongside substantial unique content. Document-level deduplication therefore struggles to remove local redundancy without discarding valuable document-specific content.

To address this limitation, several data processing pipelines adopt subdocument deduplication, which identifies and removes repeated regions within documents at finer granularities, such as paragraphs, sentences, lines, or substrings~\citep{wenzek2019ccnet,penedo2024fineweb,gohari2026gneissweb}. Prior work has shown that removing repeated substrings can reduce verbatim memorization and improve training efficiency~\citep{lee2022deduplicating}. Subdocument deduplication raises three central questions:
(1) \emph{how duplicate content should be defined and detected},
(2) \emph{how detection can be scaled to large-scale corpus}, and
(3) \emph{how many copies of each detected duplicate should be retained}.

The definition of duplication depends on the detection unit and matching criterion. Suffix arrays can identify variable-length exact repetitions without relying on predefined segmentation boundaries, whereas segmentation-based methods first divide documents into natural units and then match these units using exact or approximate signals. Although suffix arrays provide flexible substring-level detection, constructing and querying a single global suffix array is costly at corpus scale. Practical implementations therefore commonly partition the corpus into shards and perform matching independently within each shard~\citep{lee2022deduplicating,penedo2023refinedweb,gohari2026gneissweb}. Because copies assigned to different shards are never compared, shard-local matching observes only the within-shard portion of each global duplicate group and does not directly reveal its global frequency. Consequently, the deduplication result is sensitive to both the number and composition of shards, which determine which duplicate copies are placed together and can therefore be detected. In contrast, segmentation-based hashing naturally maps to distributed key--value aggregation, allowing identical units to be counted globally regardless of their physical partitioning. Its main limitation is that repetitions remain undetected when they either cross predefined segmentation boundaries or occur as substrings within a segment.

Even after duplicate groups have been identified, deciding how many copies to retain remains a critical issue. In sharded suffix-array pipelines, copy retention is implicitly induced by shard-local co-occurrence rather than explicitly controlled according to the global frequency of each duplicate group. Although hash-based aggregation can directly compute global duplicate frequencies, existing pipelines often apply frequency-independent retention policies, such as keep-one or keep-$k$~\citep{wenzek2019ccnet,penedo2024fineweb}. Such policies cannot adapt to the heterogeneous repetition patterns found in real-world corpora. High-frequency repeated segments are more likely to correspond to webpage templates, automatically generated content, or other low-information-density structures and may therefore warrant stronger compression~\citep{kohlschuetter2010boilerplate,wenzek2019ccnet}. In contrast, low-frequency repetitions may arise from reasonable quotations, version evolution, or localized content reuse and may still contain useful information~\citep{muennighoff2023scaling}. A suitable retention policy should therefore adapt the retained copy budget to the frequency and length of each duplicate group rather than assigning the same budget uniformly.

To address these limitations, we propose a scalable subdocument deduplication framework that explicitly decouples duplicate detection from copy retention. For duplicate detection, we segment documents at natural boundaries and use normalized exact hashing with distributed aggregation, allowing all occurrences of the same duplicate unit to be counted globally regardless of shard placement. For copy retention, we introduce an explicit retention function that adaptively allocates the copy budget according to the global frequency and text length of each duplicate group. The resulting policy treats low-frequency repetitions conservatively, applies stronger deletion to high-frequency repetitions, and further reduces the retention budget for long repeated spans that are more likely to reflect template reuse or direct copying. Coherence-preserving deletion is then applied to reduce document fragmentation.
Experiments on FineWeb-Edu and a code-containing web corpus show that our method achieves the best overall performance on FineWeb-Edu and consistent gains across all evaluated benchmarks on the code-containing web corpus.

Our contributions are as follows:
\begin{itemize}[leftmargin=1.4em]
\item We propose a scalable subdocument deduplication framework that decouples duplicate detection from copy retention. Natural-boundary segmentation, normalized exact hashing, and distributed aggregation enable global duplicate counting without making duplicate visibility dependent on shard placement.
\item We introduce an explicit frequency- and length-aware copy-retention function that adaptively assigns a retention budget to each duplicate group, applying stronger compression to high-frequency or long repetitions while treating low-frequency repetitions more conservatively.
\item We evaluate the framework on FineWeb-Edu and a code-containing web corpus. Models trained on the resulting data achieve the best overall performance on the FineWeb-Edu evaluation suite and consistent gains across all evaluated benchmarks in the code-containing web setting.
\end{itemize}

\section{Related Work}
\label{sec:related-work}

\subsection{Subdocument Duplicate Detection}

Existing methods for subdocument duplicate detection mainly differ in their detection units and matching criteria. Suffix-array-based methods identify variable-length exact repeated substrings without requiring predefined segmentation boundaries~\citep{manber1993suffix}. Lee et al.~\citep{lee2022deduplicating} develop a scalable exact-substring deduplication method based on suffix arrays to detect and remove repeated spans above a predefined length threshold. RefinedWeb~\citep{penedo2023refinedweb} applies exact-substring deduplication in its web-scale processing pipeline, while GneissWeb~\citep{gohari2026gneissweb} extends this line of work with sharded exact-substring deduplication. Segmentation-based methods instead divide documents into predefined natural units and compare their exact or normalized representations. CCNet~\citep{wenzek2019ccnet}, for example, performs paragraph-level exact deduplication. DCLM~\citep{li2024dclm} uses Bloom-filter-based n-gram matching for document- and paragraph-level deduplication. Suffix-array methods offer flexible matching of variable-length spans, whereas predefined-unit and n-gram-based methods are easier to distribute but depend on the selected units or overlap criterion.

\subsection{Copy-Retention Strategies}

Once duplicate content is detected, existing pipelines generally apply fixed deletion or retention rules. In web-scale exact-substring pipelines, matching may be performed independently within shards, such that only occurrences assigned to the same shard can be compared and removed~\citep{gohari2026gneissweb}. The resulting number of retained copies therefore depends implicitly on the corpus partitioning. Segmentation-based pipelines more commonly use explicit but fixed policies, such as retaining one occurrence per duplicate group or filtering units whose frequencies exceed a predefined threshold~\citep{wenzek2019ccnet,grattafiori2024llama3}. FineWeb~\citep{penedo2024fineweb} evaluates several global line-level keep-one variants and reports that they underperform its per-snapshot MinHash-deduplicated baseline. Other work controls the contribution of repetitive content without directly deleting all detected copies. SoftDedup~\citep{he2024softdedup} downweights documents according to their estimated commonness, while FineWeb2~\citep{penedo2025fineweb2} uses duplicate-cluster information to construct deduplication-aware resampling strategies.

\section{Preliminaries}

\label{sec:motivation}

In this section, we first analyze the frequency, length, and content characteristics of duplicate chunks to clarify the basic requirements that a copy retention strategy should satisfy. We then examine the implicit copy retention behavior induced by sharded suffix array deduplication.

\subsection{Characteristics of Duplicate Chunks}
\label{sec:motivation-characteristics}

\begin{figure*}[t]
\centering
\includegraphics[width=0.95\textwidth]
{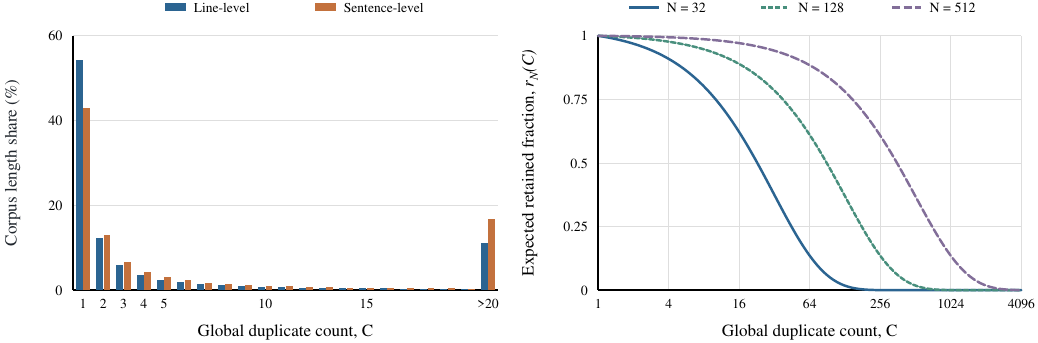}
\caption{
The left panel shows the length-weighted distributions of global
duplicate frequency for line- and sentence-level chunks after
document-level MinHash deduplication. Frequencies from 1 to 20 are shown
individually, while the final bucket aggregates chunks with $C(z)>20$.
The right panel shows the expected retained fraction $r_N(C)$ under
singleton-only shard-local retention for different effective shard
counts $N$.
}
\label{fig:frequency_distribution}
\end{figure*}

\paragraph{Frequency and content.}
After applying global document-level MinHash deduplication, we segment
the resulting corpus into line- and sentence-level chunks. We then
normalize the chunks and compute the global occurrence frequency of each
normalized chunk across the corpus. For a normalized chunk $z$, let
$C(z)$ denote its global frequency. The left panel of
Figure~\ref{fig:frequency_distribution} reports the share of analyzed
text length contributed by chunks in each frequency bucket. Frequencies
from 1 to 20 are shown individually, while the final bucket aggregates
all chunks with $C(z)>20$.

The distributions exhibit pronounced long tails at both segmentation
granularities. Unique chunks with $C(z)=1$ account for 54.2\% and
42.9\% of the analyzed text length at the line and sentence levels,
respectively. Low-frequency duplicate chunks with
$C(z)\in\{2,3,4\}$ contribute a further 21.8\% and 23.9\%. At the
other extreme, chunks occurring more than 20 times still account for
11.1\% and 16.7\% of the analyzed text length. These results show that
the corpus contains both substantial low-frequency repetition and a
non-negligible high-frequency tail, motivating a retention policy that
treats different frequency regimes differently.

To examine the relationship between frequency and content type, we
stratify duplicate chunks by frequency and manually inspect
representative duplicate groups from each stratum. High-frequency
duplicates predominantly consist of webpage templates, navigation
elements, copyright and license statements, advertisements,
automatically generated messages, and other site-level boilerplate.
In contrast, low-frequency duplicates exhibit more diverse content,
including ordinary natural-language paragraphs, reasonable quotations,
text shared across different versions, and localized reposted content.
High-frequency duplicates are therefore more likely to represent
templated and highly redundant content, for which retaining additional
copies provides little marginal information. Low-frequency duplicates,
by contrast, are more likely to contain information worth retaining.

\paragraph{Length and content.}
We further perform stratified sampling and content inspection on duplicate chunks of varying lengths within the same or similar frequency ranges. Results show that longer duplicate chunks more often correspond to complete webpage templates, fixed disclaimers, or large copied passages; shorter duplicate chunks are more heterogeneous in content, including titles, common phrases, and other short expressions.

This distinction suggests that, after controlling for frequency, chunk length still provides additional information regarding the degree of content redundancy. When long chunks are repeated identically across multiple positions, they are more likely to arise from full-template reuse or direct copying, and their different copies typically have low marginal information. In contrast, short text segments may appear independently in different contexts; identical surface forms do not necessarily imply actual copying relationships, and overly aggressive deletion may inadvertently remove legitimate linguistic reuse. Therefore, for duplicate chunks with the same frequency, the retention strategy should apply stronger compression to longer chunks.

These observations suggest that retention should depend on both frequency and length. Frequency captures how widely a chunk is repeated, while length helps distinguish likely copied or templated content from short expressions that may recur naturally.

\subsection{Implicit Retention Behavior of Shard-Local Deduplication}
\label{sec:motivation-sharding}

The foregoing analysis suggests that a copy-retention strategy should treat low-frequency duplicates conservatively while applying stronger compression to high-frequency duplicates. However, directly specifying a function with this behavior would be largely ad hoc. We observe that random shard-local deduplication naturally induces a similar frequency-dependent retention pattern: low-frequency duplicate groups are more likely to appear as isolated copies within shards, whereas high-frequency groups are more likely to contain multiple co-located copies and undergo stronger deletion. We therefore analyze this implicit retention behavior and use it to derive an explicit retention function.

At corpus scale, suffix-array deduplication is typically performed independently across multiple shards because constructing and querying a single global suffix array is computationally expensive~\citep{lee2022deduplicating,penedo2023refinedweb,gohari2026gneissweb}. Because only copies assigned to the same shard are compared, whether a copy is retained depends on the number of matching copies within its shard rather than directly on the global occurrence count. Under the shard-local deletion abstraction considered here, a copy is retained only when it is the sole occurrence of its duplicate group within a shard; if two or more copies occur in the same shard, the corresponding repeated regions are removed. We refer to this behavior as \emph{singleton-only retention}.

To isolate the effect of shard count, we analyze an idealized random-sharding model. Suppose that a duplicate group contains \(C\) copies in the global corpus, where \(C\in\mathbb{N}_{+}\), and that each copy is assigned independently and uniformly at random to one of \(N\) shards, where \(N\in\mathbb{N}\) and \(N\geq 2\). Let \(X_i\) denote the number of copies assigned to the \(i\)-th shard. The shard-level copy counts jointly follow a multinomial distribution,
$
(X_1,\ldots,X_N)
\sim
\operatorname{x}
\left(
C;\frac{1}{N},\ldots,\frac{1}{N}
\right),
$
and the marginal distribution of each \(X_i\) is
$
X_i \sim \operatorname{Binomial}\left(C,\frac{1}{N}\right).
$
Under singleton-only retention, the \(i\)-th shard contributes one retained copy if and only if \(X_i=1\). Let
\(R=\sum_{i=1}^{N}\mathbf{1}[X_i=1]\)
denote the total number of retained copies. Its expectation is
\begin{equation}
\label{eq:expected-retained-copies}
\begin{aligned}
g_N(C)
&= \mathbb{E}[R]
 = \sum_{i=1}^{N}\Pr(X_i=1) \\
&= N\Pr(X_1=1)
 = N\binom{C}{1}\frac{1}{N}
   \left(1-\frac{1}{N}\right)^{C-1} \\
&= C\left(1-\frac{1}{N}\right)^{C-1}.
\end{aligned}
\end{equation}
Therefore, the expected retention ratio is
\begin{equation}
\label{eq:expected-retention-ratio}
r_N(C)
= \frac{g_N(C)}{C}
= \left(1-\frac{1}{N}\right)^{C-1}.
\end{equation}

This expression characterizes the expected behavior under uniform random assignment. Source-aware or otherwise non-random shard construction may produce different retention behavior, further illustrating the dependence of shard-local deduplication on the physical partitioning scheme.

Since \(0<1-\frac{1}{N}<1\), the retention ratio \(r_N(C)\) decreases monotonically with global frequency \(C\). As shown in the right panel of Figure~\ref{fig:frequency_distribution}, low-frequency duplicate groups rarely contain multiple copies within the same shard and therefore retain most of their copies. As \(C\) increases, intra-shard co-occurrence becomes more likely, and the retained fraction decreases. A larger \(N\) produces a slower decay, whereas a smaller \(N\) results in stronger compression.

The induced curve exhibits the desired frequency dependence. However, in shard-local suffix-array pipelines, its shape is coupled to the physical sharding configuration and cannot be controlled independently of data partitioning. We therefore abstract the curve from the sharding mechanism and use it as the basis of an explicit frequency-aware retention function. After global hash-based counting, \(N\) is reinterpreted as an effective hyperparameter controlling compression strength rather than as the number of physical shards. The length-aware adjustment is introduced in Section~\ref{sec:retention-function}.

\section{Method}
\label{sec:method}

\subsection{Overall Pipeline}
\label{sec:overall-pipeline}

Our framework consists of five stages: segmentation, text normalization, global duplicate counting, document reconstruction, and retention-guided deletion.

\paragraph{Segmentation.}
Given a document collection $\mathcal{D}$, we segment each document into a sequence of contiguous text units, referred to as \emph{chunks}. Initial chunks are obtained at natural boundaries, such as paragraph separators, line breaks, and sentence-ending punctuation~\citep{wenzek2019ccnet,penedo2024fineweb}. We then scan the chunks from left to right. If a chunk is shorter than the minimum segmentation threshold $\tau_{\mathrm{seg}}$, it is successively merged with the following chunks until its accumulated length reaches $\tau_{\mathrm{seg}}$. A residual chunk at the end of a document is retained even if its length remains below $\tau_{\mathrm{seg}}$. All chunk lengths in this stage are measured in characters. This procedure prevents short titles, list markers, and isolated phrases from becoming independent detection units while preserving the complete document content.

For documents containing code, we additionally preserve code structure. Markdown code chunks and brace-delimited structures with reliably matched boundaries are treated as indivisible units and are not segmented internally. This rule reduces the risk of breaking syntactic or structural integrity during segmentation, a concern that is specific to source-code corpus construction and deduplication~\citep{kocetkov2022stack}.

\paragraph{Text normalization.}
Let $\operatorname{norm}(x)$ denote the normalized text of a chunk occurrence $x$. For natural-language chunks, numeric expressions such as dates, page numbers, prices, and statistical values are replaced with unified placeholders, allowing structurally identical template variants to be grouped together~\citep{wenzek2019ccnet,penedo2024fineweb}. For code chunks, no content normalization is applied, and we set $\operatorname{norm}(x)=x$ to avoid grouping code that differs in constants or other potentially meaningful tokens. Normalization is deterministic and is applied before hashing.

Chunk occurrences with identical normalized text form a \emph{duplicate group}. Normalization is used only for matching and frequency counting; the original text and positional information of each occurrence are retained for document reconstruction and output.

\paragraph{Counting.}
For each chunk occurrence $x$, we compute a content hash from its normalized text as
$h(x)=H(\operatorname{norm}(x))$, where $H(\cdot)$ denotes a content hash function. The resulting hash is used as the key for distributed aggregation. Let $\mathcal{X}(\mathcal{D})$ denote the multiset of all chunk occurrences extracted from $\mathcal{D}$. For a normalized text representation $z$, its global frequency is defined as
\begin{equation}
\label{eq:global-frequency}
C(z)
=
\sum_{x\in\mathcal{X}(\mathcal{D})}
\mathbf{1}\!\left[
\operatorname{norm}(x)=z
\right],
\end{equation}
where $\mathbf{1}[\cdot]$ denotes the indicator function. This stage is a standard key--value aggregation operation and can be implemented using the shuffle-and-aggregate pattern in MapReduce or Spark~\citep{dean2004mapreduce,zaharia2012rdd}.

\paragraph{Reconstruction.}
After frequency counting, we join the duplicate-group metadata, including its global frequency and normalized-text length, back to the corresponding chunk records. We then reconstruct each document according to its document identifier and original chunk positions. This stage restores document order and context for subsequent deletion.

\paragraph{Deletion.}
The retention function $T(C,L)$ introduced in Section~\ref{sec:retention-function} assigns an initial retention budget to each duplicate group. For a duplicate group with frequency $C$ and chunk length $L$, we order its $C$ occurrences by document ID and retain the first $T(C,L)$ occurrences. Let $\operatorname{rank}(o)\in\{1,\ldots,C\}$ denote the position of occurrence $o$ in the resulting order. Its initial retention decision is
\begin{equation}
\label{eq:occurrence-retention-rule}
\operatorname{keep}_0(o)
=
\mathbf{1}\!\left[
\operatorname{rank}(o)\leq T(C,L)
\right].
\end{equation}
When the retention boundary falls within a group of occurrences sharing the same document ID, we retain the entire group, including those ranked immediately beyond $T(C,L)$. All remaining occurrences beyond the retention budget are marked as candidate deletion chunks. However, we do not delete these candidates independently. Instead, we group consecutive candidate deletion chunks within each document to reduce fragmentation caused by isolated local deletions. Let $(x_s,\ldots,x_t)$ denote a maximal contiguous run of candidate deletion chunks, and let $\ell(x_j)$ denote the original character length of chunk $x_j$. The entire run is deleted only if
\begin{equation}
\label{eq:deletion-threshold}
\sum_{j=s}^{t}\ell(x_j)
\geq
\tau_{\mathrm{del}}.
\end{equation}
Otherwise, all chunks in the run are retained. Because the rule is applied to maximal contiguous runs, the accumulated length is reset whenever a non-candidate chunk is encountered.

The chunk length $L$ at the duplicate-group level and the run length at the document level serve different purposes: the former adjusts the initial retention budget of a duplicate group, whereas the latter determines whether a candidate region is sufficiently long to be removed without excessive fragmentation. Consequently, $T(C,L)$ specifies only the initial retention budget, and the final number of retained occurrences may be larger after coherence-aware deletion is applied.

\subsection{Frequency- and Length-Aware Retention Function}
\label{sec:retention-function}

Based on the analysis in Section~\ref{sec:motivation}, we combine a frequency-derived base budget with a length-aware adjustment to determine the number of initially retained copies.

\paragraph{Frequency-Aware Retention Budget.}
Under the shard-based singleton-only retention behavior analyzed in Section~\ref{sec:motivation-sharding}, the expected number of retained copies for a duplicate group with global frequency $C$ is $g_N(C)
=
C\left(1-\frac{1}{N}\right)^{C-1}$,
and the corresponding retained fraction is
$r_N(C)
=
\left(1-\frac{1}{N}\right)^{C-1}$.
The retained fraction $r_N(C)$ decreases monotonically with $C$, thereby applying progressively stronger relative compression to higher-frequency duplicate groups. We use the corresponding expected retained count $g_N(C)$ as the base retention budget.

The parameter $N$ originally denotes the number of physical shards. After adopting distributed exact hash counting, we reinterpret it as an effective hyperparameter controlling the shape of the frequency-dependent retention curve. A larger $N$ causes the retained fraction to decay more slowly with frequency and therefore preserves more low- and medium-frequency repetitions, whereas a smaller $N$ applies stronger compression to medium- and high-frequency duplicate groups.

\paragraph{Length-Aware Adjustment.}
Section~\ref{sec:motivation-characteristics} shows that chunk length provides additional information about the likely redundancy of duplicate content. The base budget $g_N(C)$ depends only on global frequency and therefore assigns the same budget to duplicate groups with identical frequencies but different lengths. However, longer repeated chunks are more likely to correspond to templates, fixed disclaimers, or large copied passages and should therefore receive stronger compression.

Let $L$ denote the character length of the normalized chunk text. Because chunk lengths are unbounded and their distributions can vary substantially across corpora, we introduce a reference length $L_0>0$ that controls the decay rate and saturation point of the length adjustment.

The length adjustment function $\alpha(L)$ should take values in $[0,1]$, decrease monotonically with $L$, and remain constant once $L\geq L_0$. We use the following truncated linear function:
\begin{equation}
\label{eq:length-decay}
\alpha(L)
=
\max\!\left\{
0,\,
1-\frac{L}{L_0}
\right\}.
\end{equation}
When $0\leq L<L_0$, $\alpha(L)$ decreases linearly from $1$ to $0$, assigning higher weights to shorter chunks and lower weights to longer chunks. For $L\geq L_0$, the adjustment saturates at $\alpha(L)=0$. This form introduces only one parameter, is monotonic and interpretable, and avoids imposing an unbounded length penalty on extremely long chunks.

\paragraph{Joint Frequency--Length Retention.}
Combining the frequency-derived budget with the length adjustment, we define the initial number of retained copies for each duplicate group as
\begin{equation}
\label{eq:retention-threshold}
T(C,L)
=
\left\lceil
1+\bigl(g_N(C)-1\bigr)\alpha(L)
\right\rceil.
\end{equation}
The resulting rule has intuitive boundary behavior. When $L$ is small relative to $L_0$, $\alpha(L)$ remains close to $1$, and the retention budget is primarily determined by $g_N(C)$. For duplicate groups with $g_N(C)>1$, increasing $L$ monotonically decreases the underlying real-valued budget from $g_N(C)$ toward $1$, while the integer-valued $T(C,L)$ changes in discrete steps because of the ceiling operation. When $g_N(C)\leq1$, the construction yields $T(C,L)=1$ for all $L$, so the length adjustment has no further effect. Once $L\geq L_0$, $\alpha(L)=0$ and the initial retention budget becomes $T(C,L)=1$.

The additive anchor at $1$ guarantees that at least one occurrence is initially retained, while the ceiling operation converts the real-valued budget into a conservative integer budget and avoids reducing the retained count through downward rounding. The resulting function is a deterministic extension of the shard-based singleton-only retention curve rather than a stochastic simulation.

For any $C\geq1$ and $N>1$, the retention budget satisfies $1\leq T(C,L)\leq C$.
The parameter $L_0$ controls the decay rate and saturation point of the length adjustment. For chunks satisfying $L\geq L_0$, the adjustment reaches its maximum strength, and the duplicate group retains one initial occurrence.

\begin{table*}[t]
\centering
\setlength{\belowcaptionskip}{5pt}
\caption{
Downstream performance after 36K training steps on FineWeb-Edu
processed using different deduplication methods.
Average scores are computed over all eight benchmarks.
}
\label{tab:fineweb-main}

\small
\setlength{\tabcolsep}{4.5pt}
\renewcommand{\arraystretch}{1.18}

\resizebox{\linewidth}{!}{%
\begin{tabular}{l*{9}{c}}
\toprule
\multirow{2}{*}{\textbf{Method}}
& \multicolumn{2}{c}{\textbf{Knowledge}}
& \multicolumn{2}{c}{\textbf{Reasoning}}
& \multicolumn{2}{c}{\textbf{Mathematics}}
& \multicolumn{2}{c}{\textbf{Comprehensive}}
& \multirow{2}{*}{\textbf{Average}} \\

\cmidrule(lr){2-3}
\cmidrule(lr){4-5}
\cmidrule(lr){6-7}
\cmidrule(lr){8-9}

& \textbf{NQ}
& \textbf{TriviaQA}
& \textbf{HSwag}
& \textbf{PIQA}
& \textbf{GSM8K}
& \textbf{MATH}
& \textbf{MMLU}
& \textbf{CMMLU}
& \\
\midrule

\textsc{FineWeb-Edu}
& 25.87
& \second{78.75}
& 76.83
& \second{81.12}
& 32.17
& 10.61
& 58.67
& 51.07
& 51.89 \\

\textsc{Doc-MinHash}
& 27.65
& 77.92
& 76.07
& 80.79
& 29.74
& \second{10.71}
& 57.71
& 51.47
& 51.51 \\

\textsc{Suffix-Array}
& \best{28.67}
& 78.06
& \second{77.37}
& 81.07
& 33.51
& 10.70
& 60.03
& 52.89
& 52.79 \\

\textsc{Keep-One}
& 24.32
& 77.36
& 77.10
& \second{81.12}
& \second{34.42}
& 9.85
& \second{60.23}
& 52.68
& 52.14 \\

\midrule

\rowcolor{oursbg}
\textsc{Ours} (Subdoc-only)
& 27.01
& 78.06
& 77.03
& 80.63
& \best{36.19}
& 9.90
& \best{60.78}
& \best{53.77}
& \best{52.92} \\

\rowcolor{oursbg}
\textbf{\textsc{Ours}}
& \second{28.14}
& \best{78.89}
& \best{77.73}
& \best{81.34}
& 33.51
& \best{10.90}
& 59.43
& \second{53.24}
& \second{52.90} \\

\bottomrule
\end{tabular}%
}
\end{table*}

\section{Experiments}
\label{sec:experiments}

\subsection{Experimental Setup}
\label{sec:experimental-setup}

\paragraph{Implementation Details.}
We apply the deduplication methods to two source corpora and train Hy-MT2-30B-A3B models~\citep{zheng2026hymt2familyfastefficient} on the resulting corpora. For the general-web setting, the source corpus is FineWeb-Edu~\citep{penedo2024fineweb}, containing 6.28T tokens before additional deduplication. For the code-containing web setting, we construct a 561.53B-token source corpus by applying a code classifier to webpages from Common Crawl. All corpus sizes are measured using the Hy-MT2-30B-A3B tokenizer.
Each optimization step processes 2,000 sequences of length 4,096, corresponding to 8.192M tokens. All models are trained for 36,000 steps, consuming 294.9B tokens in total. The learning rate is linearly warmed up to a peak value of $3\times10^{-4}$ over the first 2,000 steps. We evaluate the models every 2,000 steps from step 2,000 through step 36,000.
For deduplication, we set $\tau_{\mathrm{seg}}=32$ characters, $\tau_{\mathrm{del}}=100$ characters, and $L_0=512$ characters. We consider $N\in[20,100]$ and use $N=100/3$ in the final configuration. For the document-level MinHash baseline, we use 5-gram shingles, 128 MinHash permutations, and a target Jaccard similarity of $0.7$.

\paragraph{Benchmarks and metrics.}
For the FineWeb-Edu experiment, we evaluate Natural Questions (NQ), TriviaQA, HellaSwag, PIQA, GSM8K, MATH, MMLU, and CMMLU~\citep{kwiatkowski2019naturalquestions,joshi2017triviaqa,zellers2019hellaswag,bisk2020piqa,cobbe2021gsm8k,hendrycks2021math,hendrycks2021mmlu,li2023cmmlu}. We report accuracy for individual generative and multiple-choice tasks and macro-averaged accuracy for multi-category suites. Each domain score is the unweighted mean of its two benchmarks, and the overall score is the unweighted mean across all eight benchmarks.
For the code-containing web experiment, we evaluate BigCodeBench-Full, HumanEval+, LiveCodeBench, FullStackBench-en, and ARC-Challenge~\citep{zhuo2024bigcodebench,liu2023evalplus,jain2024livecodebench,cheng2024fullstackbench,clark2018arc}. The overall score is the unweighted mean across the five benchmarks.

\paragraph{Baselines.}
We evaluate the following deduplication settings. \textsc{FineWeb-Edu} uses the original FineWeb-Edu corpus without deduplication. \textsc{Doc-MinHash} applies only document-level MinHash deduplication using resemblance sketches~\citep{broder1997resemblance}. Starting from the \textsc{Doc-MinHash} corpus, \textsc{Suffix-Array} performs shard-local exact substring deduplication following \citet{lee2022deduplicating}, while \textsc{Keep-One} uses the same segmentation, normalization, and duplicate grouping as our method but retains exactly one occurrence from each duplicate group. \textsc{Ours} (Subdoc-only) applies our frequency- and length-aware subdocument deduplication directly to FineWeb-Edu without document-level deduplication. The complete \textsc{Ours} pipeline first applies \textsc{Doc-MinHash}, followed by our subdocument deduplication procedure. For the code-containing web experiment, \textsc{No-Dedup} uses the source corpus without additional document- or subdocument-level deduplication.

\subsection{Main Results on FineWeb-Edu}
\label{sec:fineweb-results}

As shown in Table~\ref{tab:fineweb-main}, our method achieves top overall performance both with and without preceding document-level deduplication. With document-level MinHash, \textsc{Ours} obtains the best results on four benchmarks(TriviaQA, HellaSwag, PIQA, and MATHand) and ranks second overall with an average score of \(52.90\). Without document-level deduplication, \textsc{Ours} (Subdoc-only) leads on GSM8K, MMLU, and CMMLU and achieves the highest overall average of \(52.92\). The two configurations improve over FineWeb-Edu by \(1.01\) and \(1.03\) points, respectively, and are numerically \(0.11\) and \(0.13\) points above Suffix-Array. Together, they achieve the best result on seven of the eight benchmarks, indicating that the proposed subdocument deduplication procedure is effective under both pipeline configurations.

The baseline comparison separates the effects of detection granularity and copy retention. Doc-MinHash reduces the average from \(51.89\) to \(51.51\), whereas all four settings with subdocument-level deduplication outperform both FineWeb-Edu and Doc-MinHash. This comparison supports the benefit of removing redundancy at a finer granularity in the evaluated setting. More importantly, Keep-One uses the same segmentation, normalization, and duplicate grouping as our method but achieves a lower average of \(52.14\). This controlled comparison provides more direct evidence that adaptive copy retention is more effective than uniformly retaining one occurrence per duplicate group. Suffix-Array remains competitive, and the margins of our two variants over it are modest; nevertheless, both variants achieve stronger aggregate performance.

\begin{figure}[t]
\centering
\includegraphics[width=0.98\textwidth]{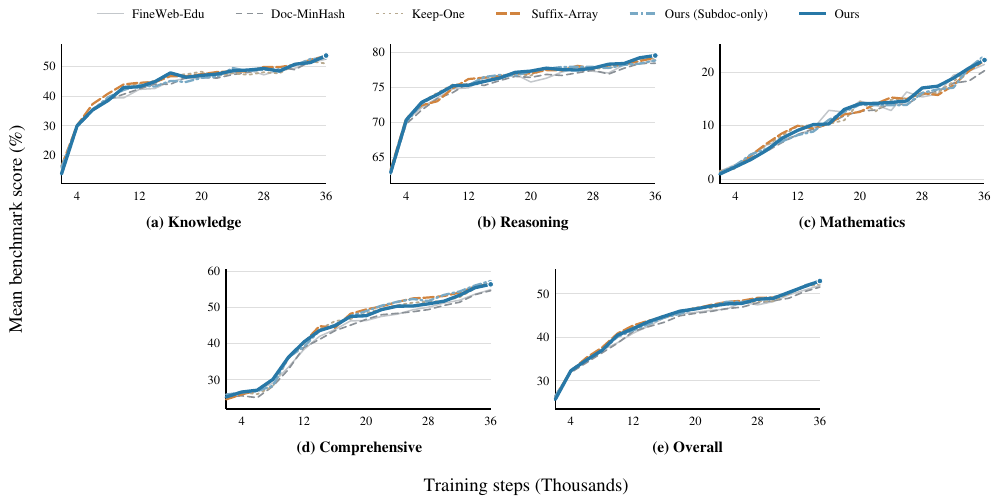}
\caption{Training dynamics of different deduplication methods, evaluated every 2k steps up to 36k. Each capability-domain panel reports the unweighted average of its two benchmarks, while Overall averages all eight benchmarks.}
\label{fig:fineweb-domain-curves}
\end{figure}

Figure~\ref{fig:fineweb-domain-curves} shows that the effects of deduplication are limited during early training, when the trajectories largely overlap. The differences become more consistent in the later stages. Both variants of our method remain near the top of the overall curve and finish above all baselines at \(36\mathrm{k}\) steps. Their domain-level trajectories differ: \textsc{Ours} ends with the highest performance in knowledge and reasoning, whereas \textsc{Ours} (Subdoc-only) performs best in mathematics and comprehensive evaluation. Suffix-Array remains competitive throughout training, consistent with its small gap in the final average. Overall, the curves suggest that the final gains reflect a sustained late-stage trend rather than an isolated fluctuation at the last checkpoint.

\subsection{Results on Code-Containing Webpages}
\label{sec:codepage-results}

\begin{table}[t]
\centering
\setlength{\belowcaptionskip}{5pt}
\caption{Performance after 36k training steps on code-containing webpages. Average denotes the unweighted mean across the five benchmarks.}
\label{tab:codepage-main}

\small
\setlength{\tabcolsep}{4.8pt}
\renewcommand{\arraystretch}{1.12}

\resizebox{\textwidth}{!}{%
\begin{tabular}{lcccccc}
\toprule
\textbf{Method}
& \textbf{BigCodeBench}
& \textbf{HumanEval+}
& \textbf{LiveCodeBench}
& \textbf{FullStackBench-en}
& \textbf{ARC-Challenge}
& \textbf{Average} \\
\midrule

\textsc{No-Dedup}
& 36.93
& 51.53
& 12.99
& 35.29
& 53.18
& 37.98 \\

\rowcolor{oursbg}
\textsc{Ours}
& \textbf{39.30}\gain{2.37}
& \textbf{57.06}\gain{5.53}
& \textbf{14.72}\gain{1.73}
& \textbf{37.37}\gain{2.08}
& \textbf{58.53}\gain{5.35}
& \textbf{41.40}\gain{3.42} \\

\bottomrule
\end{tabular}%
}
\end{table}

Table~\ref{tab:codepage-main} shows that our method improves performance on all five evaluated benchmarks, increasing the average score from \(37.98\) to \(41.40\). The largest gains are observed on HumanEval+ and ARC-Challenge, with improvements of \(5.53\) and \(5.35\) points, respectively. BigCodeBench, LiveCodeBench, and FullStackBench-en also improve by \(2.37\), \(1.73\), and \(2.08\) points. The gains across four code benchmarks indicate that structure-aware subdocument deduplication benefits several forms of code capability rather than only one specific evaluation setting. The improvement on ARC-Challenge further suggests that removing local redundancy from code-containing webpages does not trade off broader non-code capability in this experiment.

\begin{figure*}[t]
\centering
\includegraphics[width=0.98\textwidth]{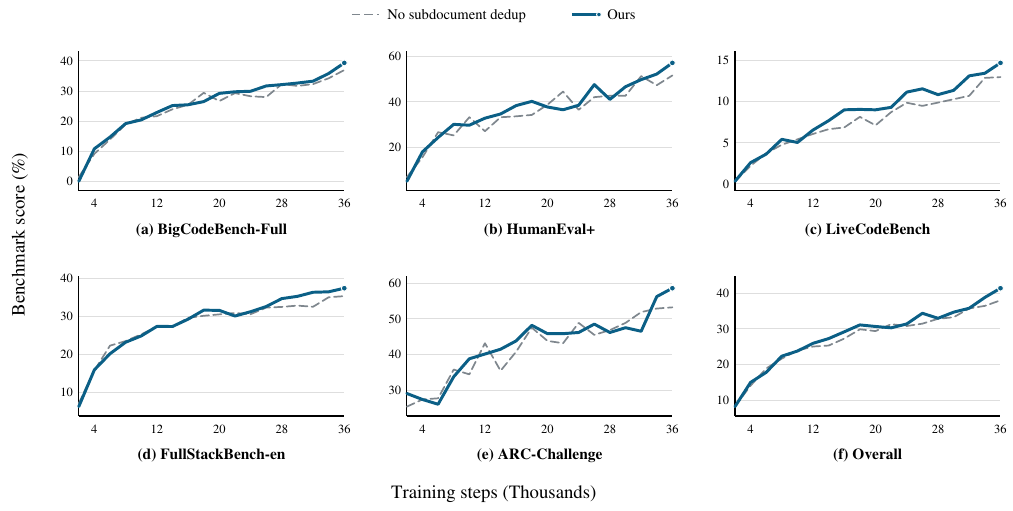}
\caption{Training dynamics on code-containing webpages. Panels (a)--(e) report the five individual benchmarks, while panel (f) shows their unweighted average. Each panel uses its own vertical scale.}
\label{fig:codepage-learning-panels}
\end{figure*}

Figure~\ref{fig:codepage-learning-panels} shows that the endpoint improvements are also reflected in the training dynamics. The two settings perform similarly during early training, after which our method develops clearer advantages on HumanEval+, LiveCodeBench, and ARC-Challenge. The overall curve follows the same pattern and remains higher toward the end of training. Therefore, the final improvement is supported by a broader late-stage trend rather than an isolated advantage at the \(36\mathrm{k}\)-step checkpoint.

\definecolor{groupbg}{gray}{0.94}

\newcolumntype{C}{>{\centering\arraybackslash}X}

\begin{table*}[t]
\centering
\small
\setlength{\belowcaptionskip}{5pt}
\caption{
Corpus sizes after document- and subdocument-level deduplication.
Retention ratios are computed relative to the corresponding raw corpus.
}
\label{tab:raw-corpus-accounting}

\setlength{\tabcolsep}{5pt}
\renewcommand{\arraystretch}{1.15}

\begin{tabularx}{\textwidth}{@{}lCCCCC@{}}
\toprule
\multirow{2}{*}{\textbf{Method}}
& \multirow{2}{*}{\textbf{Raw Tokens}}
& \multicolumn{2}{c}{\textbf{Document-Level Dedup.}}
& \multicolumn{2}{c}{\textbf{Subdocument-Level Dedup.}} \\
\cmidrule(lr){3-4}
\cmidrule(lr){5-6}
&
& \textbf{Remaining}
& \textbf{Retention}
& \textbf{Remaining}
& \textbf{Retention} \\
\midrule

\rowcolor{groupbg}
\multicolumn{6}{c}{\textbf{FineWeb-Edu}} \\

\textsc{Suffix-Array}
& 6.28T
& 1.01T
& 16.08\%
& 937.41B
& 14.93\% \\

\textsc{Keep-One}
& 6.28T
& 1.01T
& 16.08\%
& 799.55B
& 12.73\% \\

\textbf{\textsc{Ours}}
& 6.28T
& 1.01T
& 16.08\%
& 905.36B
& 14.42\% \\

\addlinespace[3pt]

\rowcolor{groupbg}
\multicolumn{6}{c}{\textbf{Code-Containing Webpages}} \\

\textbf{\textsc{Ours}}
& 561.53B
& --
& --
& 502.23B
& 89.44\% \\

\bottomrule
\end{tabularx}
\end{table*}

\subsection{Corpus Reduction Analysis}
\label{sec:corpus-reduction}

Table~\ref{tab:raw-corpus-accounting} reports the corpus sizes after each deduplication stage. On FineWeb-Edu, document-level MinHash reduces the corpus from \(6.28\)T to \(1.01\)T tokens, retaining \(16.08\%\) of the raw data. Subdocument-level deduplication further reduces the corpus to \(937.41\)B tokens for Suffix-Array, \(799.55\)B for Keep-One, and \(905.36\)B for our method, corresponding to raw-corpus retention ratios of \(14.93\%\), \(12.73\%\), and \(14.42\%\), respectively. Keep-One therefore applies the strongest additional compression, while our method retains more data than Keep-One but less than Suffix-Array.

This comparison also shows that downstream performance is not determined solely by the amount of data removed. Although Keep-One produces the smallest FineWeb-Edu corpus, it performs below both Suffix-Array and our method, whereas our method achieves the best overall benchmark performance with an intermediate retention ratio. On code-containing webpages, our method retains \(502.23\)B of the original \(561.53\)B tokens, corresponding to a retention ratio of \(89.44\%\).

\begin{figure}[H]
\centering
\includegraphics[width=0.95\textwidth]
{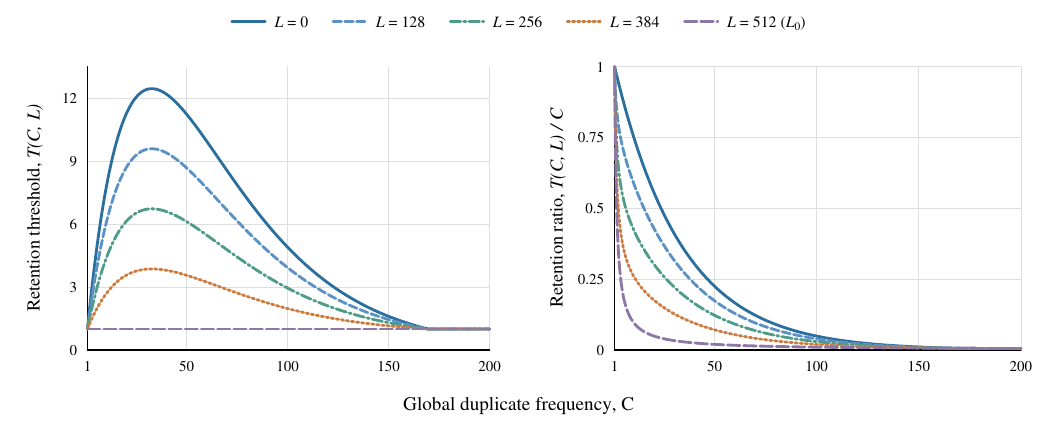}
\caption{Retention behavior under the experimental configuration
$N=100/3$ and $L_0=512$. The left shows the continuous
pre-rounding retention budget, and the right panel shows the
corresponding retained fraction for different chunk lengths. Both
panels apply the one-copy lower bound.}
\label{fig:length-aware-retention}
\end{figure}

\subsection{Retention Behavior Across Frequency and Length}
\label{sec:retention-function-behavior}

As shown in Figure~\ref{fig:length-aware-retention}, for $L<L_0$, the
absolute retention budget first increases with duplicate frequency,
reaches its maximum at approximately $C=33$, and then decreases toward
one copy. Increasing $L$ shifts the retention curve downward, so longer
repeated chunks receive smaller budgets at the same frequency. The
$L=0$ curve corresponds to frequency-only retention, whereas
$L=L_0=512$ yields a one-copy budget across all frequencies.
Although the absolute number of retained copies may initially increase,
the retained fraction decreases monotonically with duplicate frequency.
The policy therefore preserves more copies of short, low-frequency
repetitions while applying stronger compression to long or highly
frequent duplicate content.

\section{Conclusion}
\label{sec:conclusion}

In this paper, we presented a scalable subdocument deduplication framework that decouples duplicate detection from copy retention. It identifies duplicate groups through natural-boundary segmentation, normalized exact hashing, and distributed aggregation, and allocates retention budgets according to global duplicate frequency and span length. Coherence-preserving deletion further reduces fragmentation in both natural-language and code documents. Under a fixed training budget, models trained on data processed by our method achieve competitive performance. These results show that effective subdocument deduplication requires not only identifying repeated content, but also explicitly controlling how many copies to retain.

\clearpage

\bibliography{dedup}
\bibliographystyle{iclr2026_conference}

\end{document}